\documentclass[twoside]{article}
\usepackage[accepted]{aistats2018}

\usepackage{natbib}
\usepackage{amsmath}
\usepackage{amssymb}
\usepackage{hyperref}
\usepackage{url}
\usepackage{graphicx, subcaption, cleveref}

\DeclareMathOperator{\sign}{sign}
\DeclareMathOperator{\trace}{trace}
\DeclareMathOperator{\rank}{rank}

\newtheorem{theorem}{Theorem}

\begin{document}

%

%

\twocolumn[

\aistatstitle{Spectral Algorithms for Computing Fair Support Vector Machines}

\aistatsauthor{ Matt Olfat \And Anil Aswani }

\aistatsaddress{ UC Berkeley \And UC Berkeley } 


]

\begin{abstract}
Classifiers and rating scores are prone to implicitly codifying biases, which may be present in the training data, against protected classes (i.e., age, gender, or race). So it is important to understand how to design classifiers and scores that prevent discrimination in predictions.  This paper develops computationally tractable algorithms for designing accurate but fair support vector machines (SVM's).  Our approach imposes a constraint on the covariance matrices conditioned on each protected class, which leads to a nonconvex quadratic constraint in the SVM formulation.  We develop iterative algorithms to compute fair linear and kernel SVM's, which solve a sequence of relaxations constructed using a spectral decomposition of the nonconvex constraint. Its effectiveness in achieving high prediction accuracy while ensuring fairness is shown through numerical experiments on several data sets.
\end{abstract}

\section{INTRODUCTION}

The increasing prevalence of machine learning to automate decision-making systems has drawn societal scrutiny on the approaches used for verification and validation of these systems.  In particular, two main concerns have been voiced with regards to the correctness and accuracy of decisions provided.  The first is a general lack of interpretability for how predictions are produced by many learning techniques \citep{ridgeway1998interpretable,lou2012intelligible,caruana2015intelligible}, and the second is the possibility of perpetuating inequities that may be present in the training data \citep{united2014big,barocas2016big,aclu2016}.

This paper focuses on the latter: We study how to design fair support vector machines (SVM's), and our goal is to construct a classifier $h(x,t) : \mathbb{R}^p \times \mathbb{R}\rightarrow\{-1,+1\}$ that inputs predictors $x\in\mathbb{R}^p$ and a threshold $t$, and predicts a label $y\in\{-1,+1\}$, while ensuring fairness with respect to a protected class $z\in\{-1, +1\}$ (e.g., age, gender, or race).  We assume there are only two protected classes; however, our formulations generalize to the setting with multiple protected classes.


We make four main contributions. First, we reinterpret two fairness notions using receiver operating characteristic (ROC) curves, which leads to a new visualization for classifier fairness.  Second, we capture fairness by defining a constraint on covariance matrices conditioned on protected classes, which leads to a nonconvex quadratic constraint in the SVM formulation.  Third, we construct an iterative algorithm that uses a spectral decomposition of the nonconvex constraint to compute fair linear and kernel SVM's; we prove iterates converge to a local minimum.  Fourth, we conduct numerical experiments to evaluate our algorithms.

\subsection{Fairness Notions for Classifiers}

Ensuring classifiers are fair requires quantifying their fairness.  However, \cite{friedler2016possibility} and \cite{kleinberg2016inherent} showed that no single metric can capture all intuitive aspects of fairness, and so any metric must choose a specific aspect of fairness to quantify.  In this paper, we consider the arguably two most popular notions: demographic parity \citep{calders2009building,zliobaite2015relation,zafar2017} and equal opportunity \citep{dwork2012fairness,hardt2016equality}.  These notions are typically considered for a single threshold of the classifier, but here we will consider all possible thresholds.  We believe this is more in-line with malicious usage of classifiers in which strategic choice of thresholds can be used to practice discrimination.

\subsection{Algorithms to Compute Fair Classifiers}

Several approaches have been developed to construct fair classifiers.  Some \citep{zemel2013learning,louizos2015variational} compute transformations of the data to make it independent of the protected class, though this can be too conservative and reduce predictive accuracy more than desired.  Another method \citep{hardt2016equality} modifies any classifier to reduce its accuracy with respect to protected classes until fairness is achieved.  Several techniques compute a fair classifier at a single threshold \citep{calders2009building,cotter2016satisfying}; however, our interest is in classifiers that are fair at all thresholds.  The only method we are aware of that tries to compute a fair classifier for all thresholds is that of \cite{zafar2017}, which will be our main comparison.


\subsection{Outline}

After describing the data and our notation in Section \ref{sec:dan}, we next define two fairness notions and provide a new ROC visualization of fairness in Section \ref{sec:roc}.  Section \ref{sec:fc} derives constraints to improve the fairness of linear and kernel SVM's at all thresholds.  This involves nonconvex constraints, and in Section \ref{sec:sa} we present iterative algorithms that compute fair linear and kernel SVM's by solving a sequence of convex problems defined using a spectral decomposition.  Section \ref{sec:nr} conducts numerical experiments using both synthetic and real datasets to demonstrate the efficacy of our approach in computing accurate but fair SVM's.

\section{DATA AND NOTATION}
\label{sec:dan}

Our data consists of 3-tuples $(x_i, y_i, z_i)$ for $i = 1,\ldots,n$ points, where $x_i \in \mathbb{R}^p$ are predictors, $y_i \in \{-1, +1\}$ are labels, and $z_i \in \{-1,+1\}$ label a protected class.  For a matrix $W$, the $i$-th row of $W$ is denoted $W_i$.  Define $X\in\mathbb{R}^{n\times p}$, $Y\in\mathbb{R}^n$, and $Z\in\mathbb{R}^n$ to be matrices such that $X_i = x_i^\textsf{T}$, $Y_i = y_i$, and $Z_i = z_i$, respectively. 

Let $N = \{i : z_i = -1\}$ be the set of indices for protected class is negative, and similarly let $P = \{i : z_i = +1\}$ be the set of indices for which the protected class is positive.  We use $\#N$ and $\#P$ for the cardinality of the sets $N$ and $P$, respectively.  Now define $X_+$ to be a matrix whose rows are $x_i^\textsf{T}$ for $i\in P$, and similarly define $X_-$ to be a matrix whose rows are $x_i^\textsf{T}$ for $i\in N$.  Let $\Sigma_+$ and $\Sigma_-$ be the covariance matrices of $[x_i | z_i = +1]$ and $[x_i | z_i = -1]$, respectively. 

Next let $K(x,x') : \mathbb{R}^p\times\mathbb{R}^p\rightarrow\mathbb{R}$ be a kernel function, and consider the notation
\begin{equation}
\label{eqn:gram}
K(X,X') = \begin{bmatrix} K(X_1^{\vphantom{'}},X_1') & K(X_1^{\vphantom{'}},X_2') & \cdots \\ K(X_2^{\vphantom{'}},X_1') & K(X_2^{\vphantom{'}},X_2') & \cdots\\\vdots
& \vdots &\ddots\end{bmatrix}
\end{equation}
Recall that the essence of the \emph{kernel trick} is to replace $x_i^\textsf{T}x_j$ with $K(x_i,x_j)$, and so the benefit of the matrix notation given in (\ref{eqn:gram}) is that it allows us to replace $X(X')^\textsf{T}$ with $K(X,X')$ as part of the kernel trick.

Last, we define some additional notation.  Let $[n] = \{1,\ldots,n\}$, and note $\mathbf{1}(u)$ is the indicator function.
A positive semidefinite matrix $U$ is denoted $U \succeq 0$.  If $U,V$ are vectors of equal dimension, then the notation $U \circ V$ refers to their element-wise product: $(U\circ V)_i = U_i\cdot V_i$.  Also, $\mathbf{e}$ is the vector whose entries are all 1.

\section{ROC Visualization of Fairness}
\label{sec:roc}

\subsection{Demographic Parity}

One popular notion of fairness is that predictions of the label $y$ are independent of the protected class $z$.  This definition is typically stated \citep{calders2009building,zliobaite2015relation,zafar2017} in terms of a single threshold, though it can be generalized to multiple thresholds.  We say that a classifier $h(x,t)$ has demographic parity at level $\Delta$ (abbreviated as DP-$\Delta$) if
\begin{multline}
\label{eqn:ddp}
\Big|\mathbb{P}\big[h(x, t) = +1 \big| z = +1\big] - \\
\mathbb{P}\big[h(x, t) = +1 \big| z = -1\big]\Big| \leq \Delta, \ \forall t\in\mathbb{R}.
\end{multline}
To understand this, note $\mathbb{P}\big[h(x, t) = +1 \big| z = +1\big]$ is the true positive rate when predicting the protected class at threshold $t$, while $\mathbb{P}\big[h(x, t) = +1 \big| z = -1\big]$ is the false positive rate when predicting the protected class at threshold $t$.  So the intuition is that a classifier is DP-$\Delta$ if its false positive rates and true positive rates are approximately (up to $\Delta$ deviation) equal at all threshold levels for the protected class.  

\begin{figure}[h] 
\centerline{\includegraphics[width=\linewidth]{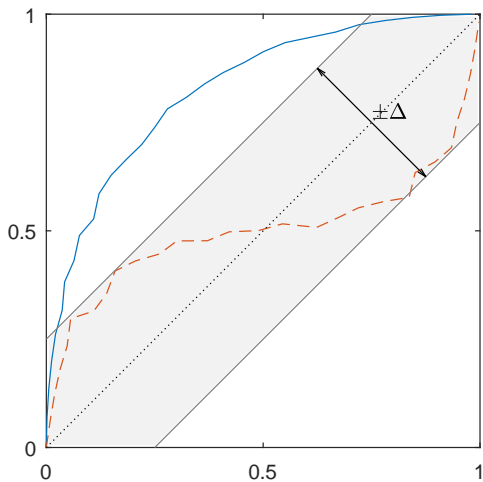}}
\caption{\label{fig: exROC} A visual representation of our notion of fairness. Here, the solid blue line is the ROC curve for the $y$ label and the dotted red line the ROC curve for the protected $z$ label. $\Delta$ refers to the maximum distance of the latter from the diagonal, which represents a perfect lack of predictability.}
\end{figure}

Reinterpreted, demographic parity requires that predictions of the classifier cannot reveal information about the protected class any better (up to $\Delta$ deviation) than random guessing.  DP-$\Delta$ is in fact equivalent to requiring that the ROC curve for the classifier $h(x,t)$ in predicting $z$ is within $\Delta$ of the \emph{line of no-discrimination}, which is the line that is achievable by biased random guessing.  More visually, Figure \ref{fig: exROC} shows how DP-$\Delta$ can be seen using an ROC curve.

\subsection{Equal Opportunity}

Demographic parity has been criticized as too strict \citep{dwork2012fairness,hardt2016equality}, and so another notion of fairness has been proposed in which predictions of the label $y$ are independent of the protected class $z$, when the true label is positive (i.e., $y = +1$).  In this definition, we must interpret $y=+1$ as a better label than $y=-1$; for instance, $y=-1$ may be a loan default, while $y=+1$ is full repayment of a loan.  This definition is typically stated \citep{hardt2016equality} in terms of a single threshold, though it can be generalized to multiple thresholds.  We say that a classifier $h(x,t)$ has equal opportunity with level $\Delta$ (abbreviated as EO-$\Delta$) if
\begin{multline}
\label{eqn:deo}
\Big|\mathbb{P}\big[h(x, t) = +1 \big| z = +1, y=+1\big] - \\
\mathbb{P}\big[h(x, t) = +1 \big| z = -1, y=+1\big]\Big| \leq \Delta, \ \forall t\in\mathbb{R}.
\end{multline}
To understand this, note $\mathbb{P}\big[h(x, t) = +1 \big| z = +1, y=+1\big]$ is the true positive rate conditioned on $y=+1$ when predicting the protected class at threshold $t$, while $\mathbb{P}\big[h(x, t) = +1 \big| z = -1, y=+1\big]$ is the false positive rate conditioned on $y=+1$ when predicting the protected class at threshold $t$.  So the intuition is that a classifier is EO-$\Delta$ if its false positive rates and true positive rates are approximately (up to $\Delta$ deviation) equal at all threshold levels for the protected class, when conditioned on $y=+1$.  

Reinterpreted, equal opportunity requires that predictions of the classifier cannot reveal information about the protected class any better (up to $\Delta$ deviation) than random guessing, when the true label is positive.  EO-$\Delta$ is equivalent to requiring that the ROC curve for the classifier $h(x,t)$ in predicting $z$ conditioned on $y=+1$ is within $\Delta$ of the line of no-discrimination.  Figure \ref{fig: exROC} shows how DP-$\Delta$ can be seen using an ROC curve.


\section{FAIRNESS CONSTRAINTS}
\label{sec:fc}
In this section, we derive several fairness constraints for linear SVM.  The kernel trick is used to convert these constraints for use in kernel SVM.  We will focus on presenting formulations for demographic parity, though all of our formulations easily generalize to equal opportunity by simply conditioning on $y=+1$. 

\subsection{Constraints for Linear SVM}

We first study constraints that can be used to ensure fairness with respect to $z$ for a linear SVM 
\begin{equation}
\label{eqn:svm}
h(x,b) = \sign(x^\textsf{T}w + b),
\end{equation}
where $w\in\mathbb{R}^p$, $b\in\mathbb{R}$ are coefficients to predict the label $y$.  Consider the following generic linear SVM formulation 
\begin{equation}
\begin{aligned}
\min\ & \textstyle \sum_{i=1}^n u_i + \lambda \|w\|_2^2 \\
\text{s.t. } & y_i^{\vphantom{\textsf{T}}}(x_i^\textsf{T} w + b) \geq 1 - u_i^{\vphantom{\textsf{T}}},&& \text{for }i\in[n]\\
& u_i \geq 0,&& \text{for }i\in[n]\\
&G(w, \theta) \leq 0.
\end{aligned}
\end{equation}
where $\lambda\in\mathbb{R}$ is a tuning parameter that can be chosen using cross-validation, and $G(w, \theta) \leq 0$ is a fairness constraint with the fairness level controlled by the (possibly vector-valued) parameter $\theta$.  We will next consider several possibilities for $G(w,\theta) \leq 0$.

\subsubsection{Indicator Constraints}

The definition of DP-$\Delta$ in (\ref{eqn:ddp}) uses probabilities, which are are not available as data.  Fortunately, we can use empirical fractions of events as an approximation:
\begin{multline}
\label{eqn:indcon}
\textstyle\Big|\frac{1}{\#P}\sum_{i \in P}\mathbf{1}(\sign(x_i^\textsf{T}w + t) = +1) - \\
\textstyle\frac{1}{\#N}\sum_{i\in N}\mathbf{1}(\sign(x_i^\textsf{T}w + t) = +1)\Big| \leq \Delta,\ \forall t\in\mathbb{R}.
\end{multline}
The above is difficult to include in the linear SVM as the fairness constraint $G(w,\theta) \leq 0$, because it involves the discontinuous and nonconvex $\sign(\cdot)$ function, and is infinite-dimensional since it must hold for all $t\in\mathbb{R}$.

\subsubsection{Integer Constraints}

An initial idea to incorporate (\ref{eqn:indcon}) in the linear SVM is based on recent empirical results \citep{miyashiro2015mixed,bertsimas2016best} where mixed integer programming (MIP) was used to exactly solve cardinality constrained linear regression.  Specifically, we can approximate the indicator constraints (\ref{eqn:indcon}) as the following mixed-integer linear inequalities
\begin{equation}
\begin{aligned}
\textstyle-\Delta\leq\frac{1}{\#P}\sum_{i \in P}v_i(t) - \frac{1}{\#N}\sum_{i\in N}v_i(t) \leq \Delta& \\
-M\cdot(1-v_i(t))\leq x_i^\textsf{T}w + t \leq M\cdot v_i(t)&\\
v_i(t) \in\{0,1\}&
\end{aligned}
\end{equation}
for $t \in \{t_1,\ldots,t_k\}$; where $M> 0$ is a large constant, and $\{t_1,\ldots,t_k\}$ is a fixed set of values. This removes the $\sign(\cdot)$ function, and it ensures a finite number of constraints.  However, we found in numerical experiments using \cite{gurobi} and \cite{mosek} that computing a fair linear SVM with the above constraints was prohibitively slow except for very small data sets.

\subsubsection{Convex Relaxation}

We next derive a convex relaxation of the indicator constraints (\ref{eqn:indcon}).  Let $M > 0$ be a constant, and consider $-M\leq u\leq M$.  Then convex upper bounds for the indicators are $\mathbf{1}(\sign(u) = +1) \leq 1+u/M$ and $-\mathbf{1}(\sign(u) = +1) \leq -u/M$.  Using these upper bounds on (\ref{eqn:indcon}) leads to the following convex relaxation 
\begin{equation}
\label{eqn:convrel}
\textstyle -d \leq \big(\frac{1}{\#P}\sum_{i \in P}x_i - \frac{1}{\#N}\sum_{i\in N}x_i\big)^\textsf{T}w \leq d,
\end{equation}
where $d = M\cdot(\Delta-1)$.  There are three important remarks about this convex relaxation.  The first is that the threshold $t$ does not appear, which means this is simply a linear constraint.  The second is that the bound $M$ can be subsumed into the parameter $d$ used to control the fairness level, meaning that this relaxation is practically independent of the bound $M$.  The third is that this convex relaxation is equivalent to the fairness constraint proposed by \cite{zafar2017}, though they derive this  using a correlation argument.

\subsubsection{Covariance Constraints}

The convex relaxation (\ref{eqn:convrel}) is fairly weak, and so it is relevant to ask whether constraints can be added to (\ref{eqn:convrel}) to increase the fairness level $\Delta$ of the resulting linear SVM.  Instead of using convex lifts \citep{lasserre2001global,gouveia2013lifts,chandrasekaran2013computational} to tighten the above constraint, we take a geometric approach to derive new constraints.

Specifically, consider the two conditional distributions $[x | z=+1]$ and $[x | z=-1]$.  Observe that the conditional distributions of $[x^\textsf{T}w + t | z=+1]$ and $[x^\textsf{T}w + t | z=-1]^\textsf{T}$ have mean $\mathbb{E}[x | z=+1]^\textsf{T}w + t$ and $\mathbb{E}[x | z=-1]^\textsf{T}w + t$, respectively.  This means (\ref{eqn:convrel}) can be interpreted as requiring
\begin{equation}
-d \leq \mathbb{E}[x^\textsf{T}w + t | z=+1] - \mathbb{E}[x^\textsf{T}w + t | z=-1]^\textsf{T}w \leq d,
\end{equation}
or equivalently that the means of $x^\textsf{T}w + t$ when conditioned on $z$ are approximately (i.e., up to $d$ apart) equal.  Thus it is natural to consider adding constraints to match the conditional distributions of $x^\textsf{T}w + t$ using higher order moments.

Here, we define a constraint to ensure that the covariances of $x^\textsf{T}w + t$ conditioned on $z$ are approximately equal.  Let $\Sigma_+$ and $\Sigma_-$ be the sample covariance matrices for $[x_i | z_i=+1]$ and $[x_i | z_i=-1]$, respectively.  Then the sample variances of $[x_i^\textsf{T}w + t | z_i = +1]$ and $[x_i^\textsf{T}w + t | z_i = -1]$ are $w^\textsf{T}\Sigma_+w$ and $w^\textsf{T}\Sigma_-w$, respectively.  So we specify our covariance constraint as
\begin{equation}
\label{eqn:covcon}
-s \leq w^\textsf{T}(\Sigma_+ - \Sigma_-)w \leq s.
\end{equation}
To our knowledge, this constraint has not been previously used to improve the fairness of classifiers.  Unfortunately, it is nonconvex because $(\Sigma_+ - \Sigma_-)$ is symmetric but typically indefinite (i.e., not positive or negative semidefinite).  Hence, computing a linear SVM with this constraint requires further development.

One obvious approach is to lift the constraint (\ref{eqn:covcon}) and then construct a semidefinite programming (SDP) relaxation \citep{goemans1995improved,luo2010semidefinite}.  Specifically, note that (\ref{eqn:covcon}) is equivalent to
\begin{equation}
\begin{aligned}
-s \leq \trace(W(\Sigma_+-\Sigma_-)) \leq s&\\
U = \begin{bmatrix} W & w \\ w^\textsf{T} & 1 \end{bmatrix} \succeq 0&\\
\rank(U) = 1&\\
\end{aligned}
\end{equation}
The above is nonconvex, but it can be convexified by dropping the $\rank(U) = 1$ constraint.  However, we found in numerical experiments using the \cite{mosek} solver that the SDP relaxation was weak and did not consistently affect the fairness or accuracy of the SVM.  Despite this result, we believe that additional convexification techniques \citep{kocuk2016inexactness,madani2017finding} can be used to strengthen the quality of the SDP relaxation; we leave the problem of how to design a strengthened SDP relaxation for future work.

\subsection{Constraints for Kernel SVM}

We next study constraints that can be used to ensure fairness with respect to $z$ for a kernel SVM 
\begin{equation}
\label{eqn:ksvm}
h(x,b) = \sign(K(X,x)^\textsf{T}(Y\circ \alpha) + b),
\end{equation}
where $\alpha\in\mathbb{R}^n$ are coefficients to predict the label $y$, and $b = \frac{1}{\#I}\sum_{i\in I}(y_i - K(X,x_i)^\textsf{T}(Y\circ \alpha))$ with the set of indices $I = \{i : 0 < \alpha_i < \lambda\}$.  Consider the following generic kernel SVM formulation 
\begin{equation}
\begin{aligned}
\min\ & \textstyle (Y\cdot\alpha)^\textsf{T}K(X,X)(Y\cdot \alpha) - \sum_{i=1}^n\alpha_i \\
\text{s.t. } & Y^\textsf{T}\alpha = 0\\
& 0 \leq \alpha_i \leq \lambda,\qquad \text{for }i\in[n]\\
&H(w, \theta) \leq 0.
\end{aligned}
\end{equation}
where $\lambda\in\mathbb{R}$ is a tuning parameter that can be chosen using cross-validation, and $H(w, \theta) \leq 0$ is a fairness constraint with the fairness level controlled by the (possibly vector-valued) parameter $\theta$.  We will next consider several possibilities for $H(w,\theta) \leq 0$.

\subsubsection{Indicator and Integer Constraints}

The indicator constraints (\ref{eqn:indcon}) can be rewritten for kernel SVM by replacing (\ref{eqn:svm}) with (\ref{eqn:ksvm}).  Unfortunately, the indicator constraints are still impractical because they are infinite-dimensional and contain a discontinuous, nonconvex function.  However, we can approximate the indicator constraints for kernel SVM as the following mixed-integer linear inequalities
\begin{equation}
\begin{aligned}
\textstyle-\Delta\leq\frac{1}{\#P}\sum_{i \in P}v_i(t) - \frac{1}{\#N}\sum_{i\in N}v_i(t) \leq \Delta& \\
-M\cdot(1-v_i(t))\leq K(X,x)^\textsf{T}(Y\circ \alpha) + t \leq M\cdot v_i(t)&\\
v_i(t) \in\{0,1\}&
\end{aligned}
\end{equation}
for $t \in \{t_1,\ldots,t_k\}$; where $M> 0$ is a large constant, and $\{t_1,\ldots,t_k\}$ is a fixed set of values. However, we found in numerical experiments using the \cite{gurobi} and \cite{mosek} solvers that computing a fair kernel SVM with the above constraints was prohibitively slow except for very small data sets.

\subsubsection{Convex Relaxation}

We next provide a convex relaxation of the indicator constraints for kernel SVM.  A similar argument as the one used for linear SVM gives
\begin{multline}
\label{eqn:kconvrel}
\textstyle -d \leq \frac{1}{\#P}\sum_{i \in P}K(X,x_i)^\textsf{T}(Y\circ \alpha) + \\
\textstyle-\frac{1}{\#N}\sum_{i\in N}K(X,x_i)^\textsf{T}(Y\circ \alpha) \leq d.
\end{multline}
Note that threshold $t$ does not appear, which means this is simply a linear constraint on $\alpha$.

\subsubsection{Covariance Constraints}

Our covariance constraint can be rewritten for the kernel SVM by first recalling that the kernel SVM with $K(x,x') = x^\textsf{T}x'$ generates the same classifier as directly solving a linear SVM.  Thus, we have the relationship $w = X^\textsf{T}(Y\circ \alpha)$ in this special case.  Next, observe that
\begin{equation}
\begin{aligned}
&\textstyle\Sigma_+ = \frac{1}{\#P}X_+^{\vphantom{\textsf{T}}}\big(\mathbb{I} - \frac{1}{\#P}\mathbf{e}\mathbf{e}^\textsf{T}\big)X_+^\textsf{T}\\
&\textstyle\Sigma_- = \frac{1}{\#N}X_-^{\vphantom{\textsf{T}}}\big(\mathbb{I} - \frac{1}{\#N}\mathbf{e}\mathbf{e}^\textsf{T}\big)X_-^\textsf{T}
\end{aligned}
\end{equation}
So if apply the kernel trick to our covariance constraint (\ref{eqn:covcon}) with the above relationships, then the resulting covariance constraint for kernel SVM becomes
\begin{equation}
\label{eqn:kcovcon}
-s \leq (Y\circ\alpha)^\textsf{T}(S_+ - S_-)(Y\circ\alpha) \leq s,
\end{equation}
where we have
\begin{equation}
\begin{aligned}
&\textstyle S_+ = \frac{1}{\#P}K(X,X_+)\big(\mathbb{I} - \frac{1}{\#P}\mathbf{e}\mathbf{e}^\textsf{T}\big)K(X,X_+)^\textsf{T}\\
&\textstyle S_- = \frac{1}{\#N}K(X,X_-)\big(\mathbb{I} - \frac{1}{\#N}\mathbf{e}\mathbf{e}^\textsf{T}\big)K(X,X_-)^\textsf{T}
\end{aligned}
\end{equation}
The constraint (\ref{eqn:kcovcon}) is a nonconvex quadratic constraint because $(S_+ - S_-)$ is symmetric but typically indefinite (i.e., not positive or negative semidefinite).

We can also construct an SDP relaxation of (\ref{eqn:kcovcon}).  Specifically, note that (\ref{eqn:kcovcon}) is equivalent to
\begin{equation}
\begin{aligned}
-s \leq \trace(W(S_+-S_-)) \leq s&\\
U = \begin{bmatrix} W & (Y\circ\alpha) \\ (Y\circ\alpha)^\textsf{T} & 1 \end{bmatrix}\succeq 0&\\
\rank(U) = 1&\\
\end{aligned}
\end{equation}
The above is nonconvex, but it can be convexified by dropping the $\rank(U) = 1$ constraint.  However, our numerical experiments using the \cite{mosek} solver found the SDP relaxation was weak and did not consistently affect the fairness or accuracy of the SVM.

\section{SPECTRAL ALGORITHM}

\label{sec:sa}

\begin{figure*}[ht]
	\centerline{\includegraphics[width=\linewidth]{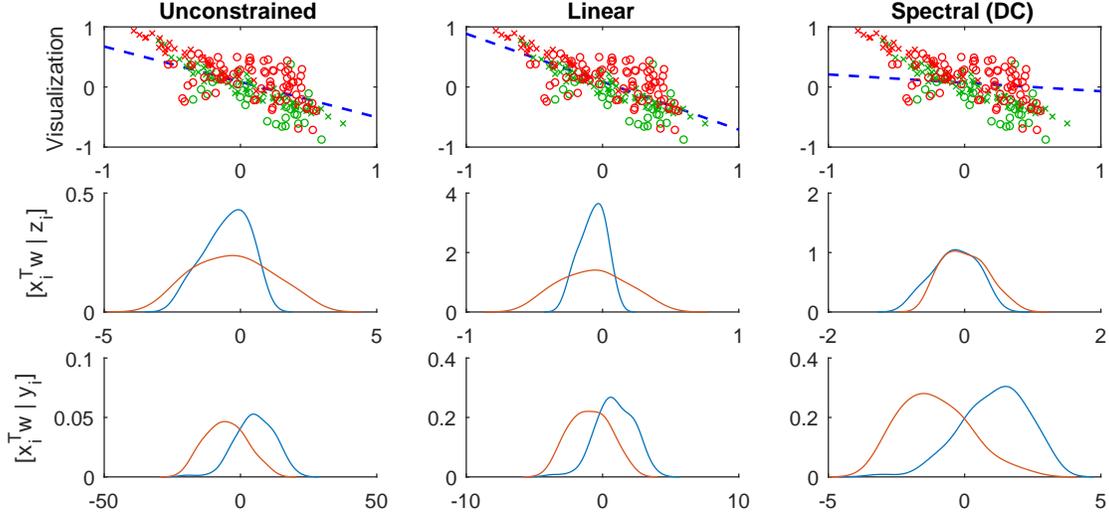}}
	\caption{\label{fig: synth} A comparison of the unconstrained, linear and spectral SVM methodologies on two-dimensional data. The first row visualizes the data, as well as the optimal support vectors from each of the methodologies. The second and third rows show the density of $x_i^\textsf{T}w$ conditioned on the $z$ and $y$ variables, respectively. In each case, the blue curve represents the density for $z_i=1$ ($y_i=1$) and the red curve the density for $z_i=-1$ ($y_i=-1$).}
\end{figure*}


The covariance constraints are conceptually promising, but they result in nonconvex optimization problems.  Here, we describe an iterative algorithm to effectively solve the SVM with our covariance constraints.

\subsection{Linear SVM}

Though we could compute the linear SVM with the covariance constraint (\ref{eqn:covcon}) using successive linearization, a better approach is possible through careful design of the algorithm: Our key observation regarding (\ref{eqn:covcon}) is that $(\Sigma_+-\Sigma_-)$ is symmetric, which means it can be diagonalized by an orthogonal matrix:
\begin{equation}
\textstyle\Sigma_+ - \Sigma_- = \sum_{i=1}^p \zeta_i^{\vphantom{\textsf{T}}} v_i^{\vphantom{\textsf{T}}}v_i^\textsf{T},
\end{equation}
where $\zeta_i\in\mathbb{R}$ and the $v_i$ form an orthonormal basis.  Now let $I_{\zeta+} = \{i : \zeta_i > 0\}$ and $I_{\zeta-} = \{i : \zeta_i < 0\}$, and define the positive semidefinite matrices
\begin{equation}
\begin{aligned}
&\textstyle U_{\zeta+} = \hphantom{-}\sum_{i\in I_{\zeta+}}\zeta_i^{\vphantom{\textsf{T}}} v_i^{\vphantom{\textsf{T}}}v_i^\textsf{T}\\ 
&\textstyle U_{\zeta-} = -\sum_{i\in I_{\zeta-}}\zeta_i^{\vphantom{\textsf{T}}} v_i^{\vphantom{\textsf{T}}}v_i^\textsf{T}
\end{aligned}
\end{equation}
This means that the function 
\begin{equation}
w^\textsf{T}(\Sigma_+ - \Sigma_-)w = w^\textsf{T}U_{\zeta+}w - w^\textsf{T}U_{\zeta-}w
\end{equation}
in the covariance constraint (\ref{eqn:covcon}) is the difference of the two convex functions $w^\textsf{T}U_{\zeta+}w$ and $w^\textsf{T}U_{\zeta-}w$.  

There is an important point to note regarding the practical importance of the spectral decomposition we performed above.  The function in the covariance constraint (\ref{eqn:covcon}) can alternatively be written as the difference of the two convex functions $w^\textsf{T}\Sigma_+w$ and $w^\textsf{T}\Sigma_-w$.  However, using this alternative decomposition yields an algorithm where the convexified subproblems are weaker relaxations than the convex subproblems generated using the spectral decomposition.  As a result, our algorithm given below is ultimately more effective because it employs the spectral decomposition.

Consequently, the constrained convex-concave procedure  \citep{tuy1995dc,yuille2002concave,smola2005kernel} can be used to design an algorithm for our setting.  We opt to use a penalized form of the quadratic constraint in our spectral algorithm to ensure feasibility always holds.  Let $w_k\in\mathbb{R}^p$ be a fixed point, and consider the optimization problem where the concave terms are linearized:
\begin{equation}
\label{eqn:ccp}
\begin{aligned}
\min\ & \textstyle \sum_{i=1}^n u_i + \lambda \|w\|_2^2 + \mu\cdot t\\
\text{s.t. } & y_i^{\vphantom{\textsf{T}}}(x_i^\textsf{T} w + b) \geq 1 - u_i^{\vphantom{\textsf{T}}},\qquad \text{for }i\in[n]\\
& u_i \geq 0,\hspace{2.94cm} \text{for }i\in[n]\\
&\textstyle -d \leq \big(\frac{1}{\#P}\sum_{i \in P}x_i - \frac{1}{\#N}\sum_{i\in N}x_i\big)^\textsf{T}w \leq d\\
&w^\textsf{T}U_{\zeta+}w - w_k^\textsf{T}U_{\zeta-}w_k - 2w_k^\textsf{T}U_{\zeta-}^\textsf{T}(w-w_k) \leq t\\
&w^\textsf{T}U_{\zeta-}w - w_k^\textsf{T}U_{\zeta+}w_k - 2w_k^\textsf{T}U_{\zeta+}^\textsf{T}(w-w_k) \leq t
\end{aligned}
\end{equation}
Our spectral algorithm for computing a fair linear SVM consists of the constrained CCP adapted to the problem of computing a linear SVM with the linear constraint (\ref{eqn:convrel}) and covariance constraint (\ref{eqn:covcon}): We initialize $w_0$ by solving a linear SVM with only the linear constraint (\ref{eqn:convrel}), and then computing successive $w_k$ by solving (\ref{eqn:ccp}).  This produces a local minimum.

\begin{theorem}[\cite{smola2005kernel}]
The spectral algorithm defined above for computing a fair linear SVM gives iterates $w_k$ that converge to a local minimum.
\end{theorem}

This theorem is simply an application of a theorem by \cite{smola2005kernel}, and the constraint qualification required by this theorem trivially holds in our case because all of our convex constraints are linear.

\subsection{Kernel SVM}

We can also design a spectral algorithm to compute fair kernel SVM's.  Since $(S_+-S_-)$ in (\ref{eqn:kcovcon}) is symmetric, t can be diagonalized by an orthogonal matrix:
\begin{equation}
\textstyle S_+ - S_- = \sum_{i=1}^p \xi_i^{\vphantom{\textsf{T}}} \nu_i^{\vphantom{\textsf{T}}}\nu_i^\textsf{T},
\end{equation}
where $\xi_i\in\mathbb{R}$ and the $\nu_i$ form an orthonormal basis.  Now let $I_{\xi+} = \{i : \xi_i > 0\}$ and $I_{\xi-} = \{i : \xi_i < 0\}$, and define the positive semidefinite matrices
\begin{equation}
\begin{aligned}
&\textstyle U_{\xi+} = \hphantom{-}\sum_{i\in I_{\xi+}}\xi_i^{\vphantom{\textsf{T}}} \nu_i^{\vphantom{\textsf{T}}}\nu_i^\textsf{T}\\ 
&\textstyle U_{\xi-} = -\sum_{i\in I_{\xi-}}\xi_i^{\vphantom{\textsf{T}}} \nu_i^{\vphantom{\textsf{T}}}\nu_i^\textsf{T}
\end{aligned}
\end{equation}
This means that the function 
\begin{multline}
(Y\circ\alpha)^\textsf{T}(\Sigma_+ - \Sigma_-)(Y\circ\alpha) = \\
(Y\circ\alpha)^\textsf{T}U_{\zeta+}(Y\circ\alpha) - (Y\circ\alpha)^\textsf{T}U_{\zeta-}(Y\circ\alpha)
\end{multline}
in (\ref{eqn:kcovcon}) is the difference of the two convex functions $(Y\circ\alpha)^\textsf{T}U_{\xi+}(Y\circ\alpha)$ and $(Y\circ\alpha)^\textsf{T}U_{\xi-}(Y\circ\alpha)$.  

Thus, the constrained convex-concave procedure  \citep{tuy1995dc,yuille2002concave,smola2005kernel} can be used to design an algorithm.  We use a penalized form of the quadratic constraint in our spectral algorithm to ensure feasibility always holds.  Let $\alpha_k\in\mathbb{R}^n$ be a fixed point, and consider the optimization problem where the concave terms are linearized:
\begin{equation}
\begin{aligned}
\min\ & \textstyle (Y\cdot\alpha)^\textsf{T}K(X,X)(Y\cdot \alpha) - \sum_{i=1}^n\alpha_i + \mu\cdot t \\
\text{s.t. } & Y^\textsf{T}\alpha = 0\\
& 0 \leq \alpha_i \leq \lambda,\qquad \text{for }i\in[n]\\
&\textstyle -d \leq \frac{1}{\#P}\sum_{i \in P}K(X,x_i)^\textsf{T}(Y\circ \alpha) + \\
&\hspace{1.75cm}\textstyle-\frac{1}{\#N}\sum_{i\in N}K(X,x_i)^\textsf{T}(Y\circ \alpha) \leq d\\
&(Y\circ \alpha)^\textsf{T}U_{\xi+}(Y\circ \alpha) - (Y\circ \alpha_k)^\textsf{T}U_{\xi-}(Y\circ \alpha_k) +\\
&\hspace{1.6cm}- 2(Y\circ \alpha_k)^\textsf{T}U_{\zeta-}^\textsf{T}(Y\circ (\alpha-\alpha_k)) \leq t\\
&(Y\circ \alpha)^\textsf{T}U_{\xi-}(Y\circ \alpha) - (Y\circ \alpha_k)^\textsf{T}U_{\xi+}(Y\circ \alpha_k) +\\
&\hspace{1.6cm}- 2(Y\circ \alpha_k)^\textsf{T}U_{\zeta+}^\textsf{T}(Y\circ (\alpha-\alpha_k)) \leq t\\
\end{aligned}
\end{equation}
Our spectral algorithm for computing a fair kernel SVM consists of the constrained CCP adapted to the problem of computing a kernel SVM with the linear constraint (\ref{eqn:kconvrel}) and covariance constraint (\ref{eqn:kcovcon}): We initialize $w_0$ by solving a kernel SVM with only the linear constraint (\ref{eqn:kconvrel}), and then computing successive $w_k$ by solving (\ref{eqn:ccp}).  This produces a local minimum.

\begin{theorem}[\cite{smola2005kernel}]
The spectral algorithm defined above for computing a fair kernel SVM gives iterates $w_k$ that converge to a local minimum.
\end{theorem}

This theorem is simply an application of a theorem by \cite{smola2005kernel}, and the constraint qualification required by this theorem trivially holds in our case because all of our convex constraints are linear.

\section{NUMERICAL EXPERIMENTS}

\label{sec:nr}

We use synthetic and real datasets to evaluate the efficacy of our approach.  We compare linear SVM's computed using our spectral algorithm (SSVM) to a standard linear SVM (LSVM) and a linear SVM computed using the approach of \cite{zafar2017} (ZSVM), since this is the only existing approach that to our knowledge is designed to ensure fairness at all thresholds.

\subsection{Synthetic Data}

	\begin{figure}[ht]
		\centering
		\begin{subfigure}[t]{0.32\linewidth}
			\includegraphics[trim = 0 0 10 10,clip,width=\linewidth]{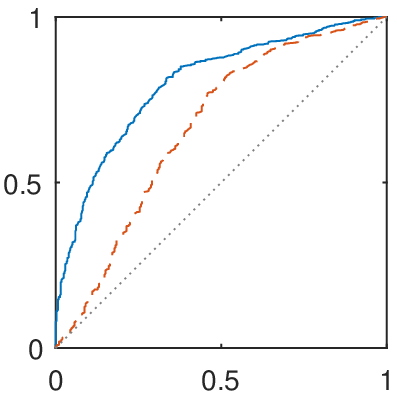}
			\caption{\label{fig: wineROCUnc}}
		\end{subfigure}
		\begin{subfigure}[t]{0.32\linewidth}
			\includegraphics[trim = 0 0 10 10,clip,width=\linewidth]{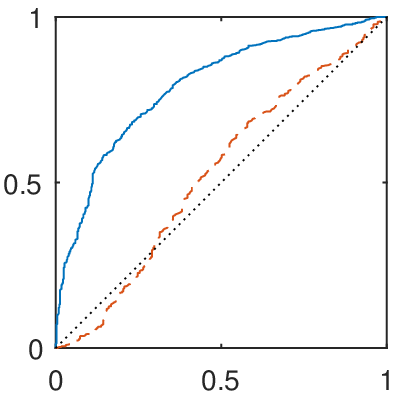}
			\caption{\label{fig: wineROCLin}}
		\end{subfigure}
        \begin{subfigure}[t]{0.32\linewidth}
			\includegraphics[trim = 0 0 10 10,clip,width=\linewidth]{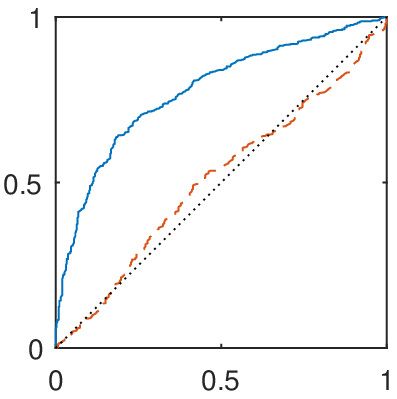}
			\caption{\label{fig: wineROCSpec}}
		\end{subfigure}
        \vspace{0.05in}
		\begin{subfigure}[t]{0.32\linewidth}
			\includegraphics[trim = 0 0 10 10,clip,width=\linewidth]{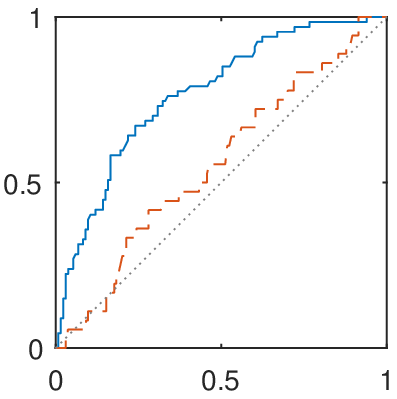}
			\caption{\label{fig: germROCUnc}}
		\end{subfigure}
        \begin{subfigure}[t]{0.32\linewidth}
			\includegraphics[trim = 0 0 10 10,clip,width=\linewidth]{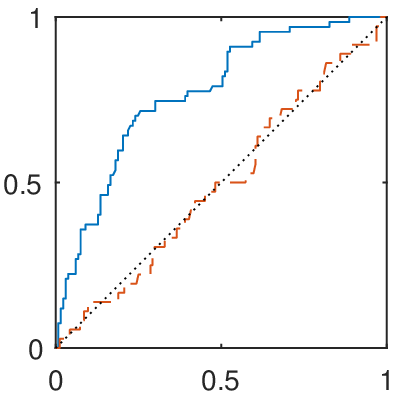}
			\caption{\label{fig: germROCLin}}
		\end{subfigure}
		\begin{subfigure}[t]{0.32\linewidth}
			\includegraphics[trim = 0 0 10 10,clip,width=\linewidth]{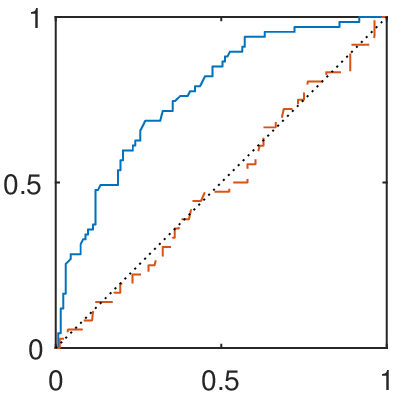}
			\caption{\label{fig: germROCSpec}}
		\end{subfigure}
		\caption{\label{fig:allroc}ROC plots for the three SVM algorithms on both datasets. In each case, the solid blue line is the ROC curve for the $y$ label and the dotted red line the ROC curve for the protected $z$ label. \Cref{fig: wineROCUnc,fig: wineROCLin,fig: wineROCSpec} show the ROC plots for LSVM, ZSVM, and SSVM on the wine quality data, and \Cref{fig: germROCUnc,fig: germROCLin,fig: germROCSpec} show the same for the German credit data.}
	\end{figure}
 
\paragraph{Experimental Design}

We seek to examine the case in which our $y$ and $z$ labels are correlated, and $X_N$ and $X_P$ have differing covariances. Thus, we generate 200 data points where the $y$ and $z$ labels are generated through logit models using two separate sets of randomly generated ``true'' parameters, with dot product between the logit parameters of $y$ and $z$ of 0.85. The singular values of the covariance matrix of $[X_i|z_i=1]$ were then skewed to generate the data seen in \Cref{fig: synth}. The empirical correlation of $y_i$ and $z_i$ is 0.45.





\paragraph{Results}

The results of the three methods using $d=0.075$ and $\mu=10$ are shown in the three columns of \Cref{fig: synth}. Here, points in $N$ and $P$ are differentiated by marker shape (``x'' and ``o'', respectively), and points with label $y_i=-1$ and $y_i=1$ are differentiated by color (red and green, respectively). If $w$ denotes the coefficients computed by each method, then the second row shows the empirical densities of $x_i^\textsf{T}w$ conditioned on the protected class $z_i$, and the third row shows the empirical densities of $x_i^\textsf{T}w$ conditioned on the label $y_i$. Fairness occurs if the conditional densities in the second row are similar, and prediction accuracy occurs if the densities in the third row are disparate.  These results show that the densities of $[x_i^\textsf{T}w | z_i=+1]$ and $[x_i^\textsf{T}w | z_i=-1]$ are distinguishable for SSVM and ZSVM, while they are almost identical for SSVM.  On the other hand, the densities of $[x_i^\textsf{T}w | y_i=+1]$ and $[x_i^\textsf{T}w | y_i=-1]$ are distinct for all three methods. 

\subsection{Real World Datasets}

\begin{figure*}[ht]
\centering
\begin{subfigure}[t]{0.45\textwidth}
	\includegraphics[width=\linewidth]{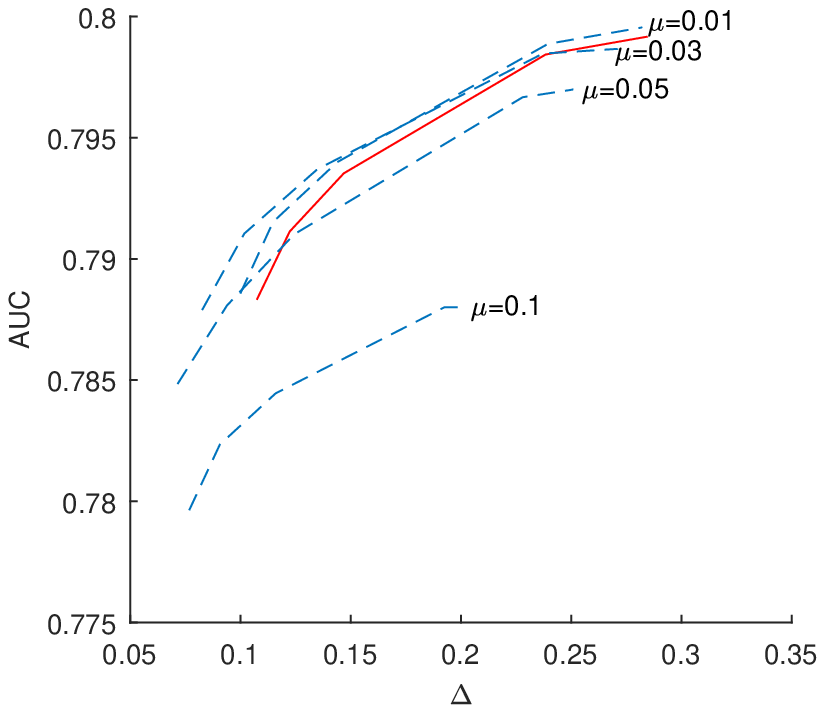}
    \caption{\label{fig: wineauc} Wine quality data.}
\end{subfigure}
\begin{subfigure}[t]{0.45\textwidth}
	\includegraphics[width=\linewidth]{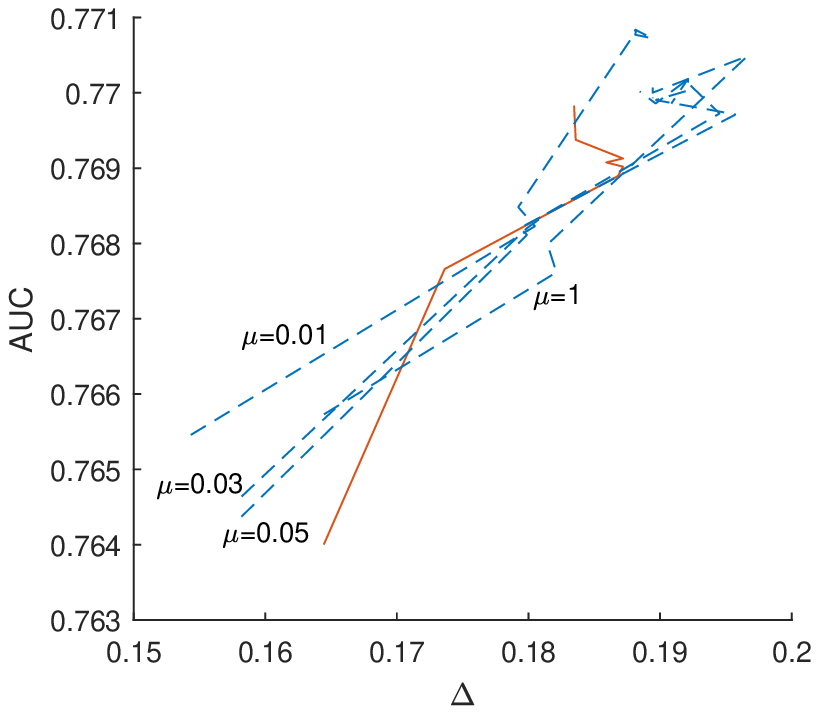}
    \caption{\label{fig: germauc} German credit data.}
\end{subfigure}
\caption{Comparing the accuracy and fairness of the ZSVM and SSM methods for various $d$ and $\mu$. The solid red line represents results for the ZSVM, and the dotted blue lines denote results for the SSVM for some $\mu$.\label{fig:allpareto}}
\end{figure*}

\paragraph{Data overview}

We next use a wine quality dataset \cite{cortez2009modeling} and a dataset of German credit card customers \citep{Lichman:2013}. The first dataset is a compilation of 12 attributes of 6,497 wines (e.g., acidity, residual sugar, alcohol content, and color), as well as a ranking out of 10 that is provided by professional taste-testers. Here, we label $y_i=1$ when a wine is rated as a 6 or above and $y_i=-1$ otherwise, and we define $z_i=1$ for white wines and $z_i=-1$ for reds. Notably, all explanatory variables are continuous. The second dataset is a compilation of 20 attributes (e.g., marriage, employment and housing status, number of existing credit lines, and age) of 1000 German applicants for loans. We label $y_i=1$ for applicants that defaulted and $y_i=-1$ for applicants that did not default, and let $z_i=1$ for applicants that are renting a home and $z_i=-1$ for applicants that own their home. Note that a large number of variables are discrete. There is no missing data in either dataset.





\paragraph{Metrics of comparison}

We compare SSVM and ZSVM based on the tradeoffs that they make between predictive accuracy for $y$, measured using the area under the ROC curve (AUC), and their fairness with respect $z$, measured by DP-$\Delta$ with respect to $z$.

\paragraph{Experimental Design}

We conducted five rounds of cross-validation on each dataset and computed the average AUC and average $\Delta$, using a 70-30 training-testing split. Within each round, we first apply $5$-fold cross-validation on LSVM to choose the $\lambda$ that maximizes AUC, and this value of $\lambda$ was used with both SSVM and ZSVM to minimize the impact of cross-validation on the comparison between methods. We varied $d$ over the values 0, 0.001, 0.002, 0.005, 0.01, 0.025, 0.05 and 0.1. And for SSVM we tried several values of $\mu$, which are shown in our plots.

\paragraph{Results}

\Cref{fig:allroc} shows representative examples of ROC curves for both datasets from one instance of cross-validation.  Both ZSVM and SSVM improve fairness with respect to LSVM while maintaining high accuracy, and SSVM ensures an even stricter level of fairness than LSVM while keeping high accuracy.  The tradeoff curves between prediction accuracy and fairness are shown in \Cref{fig:allpareto}.  Increasing $d$ generally jointly decreases fairness and increases accuracy, while small increases in $\mu$ for our SSVM can often improve both fairness and accuracy.  Large increases in $\mu$ generally increase fairness but decrease accuracy. Note that setting $\mu=0$ leads to the curve for SSVM to align with the curve of ZSVM, since they are equivalent when $\mu = 0$.  Also observe that $\Delta$ is more sensitive to changes in $d$ and $\mu$ than the AUC, which implies that we are able control fairness without losing much predictive accuracy.



\section{Conclusion}

We considered multi-threshold notions of fairness for classifiers, and designed a nonconvex constraint to improve the fairness of linear and kernel SVM's under all thresholds. We developed an iterative optimization algorithm (that uses a spectral decomposition) to handle our nonconvex constraint in the resulting problem to compute the SVM, and empirically compared our approach to standard linear SVM and an SVM with a linear fairness constraint using both synthetic and real data. We found that our method can strictly improve the fairness of classifiers for all thresholding values with little loss in accuracy; in fact, some of our results even showed a slight increase in accuracy with increasing fairness. Our work opens the door for further research in a number of areas, including hierarchies of fairness constraints considering subsequent moments of the data, and theoretical guarantees on the fairness of such classification methods.




\bibliographystyle{abbrvnat}
\bibliography{fsvm}

\begin{thebibliography}{32}
\providecommand{\natexlab}[1]{#1}
\providecommand{\url}[1]{\texttt{#1}}
\expandafter\ifx\csname urlstyle\endcsname\relax
  \providecommand{\doi}[1]{doi: #1}\else
  \providecommand{\doi}{doi: \begingroup \urlstyle{rm}\Url}\fi

\bibitem[Barocas and Selbst(2016)]{barocas2016big}
S.~Barocas and A.~D. Selbst.
\newblock Big data's disparate impact.
\newblock \emph{California Law Review}, 104, 2016.

\bibitem[Bertsimas et~al.(2016)Bertsimas, King, and
  Mazumder]{bertsimas2016best}
D.~Bertsimas, A.~King, and R.~Mazumder.
\newblock Best subset selection via a modern optimization lens.
\newblock \emph{Annals of Statistics}, 44\penalty0 (2):\penalty0 813--852,
  2016.

\bibitem[Bhandari(2016)]{aclu2016}
E.~Bhandari.
\newblock Big data can be used to violate civil rights laws, and the {FTC}
  agrees, 2016.

\bibitem[Calders et~al.(2009)Calders, Kamiran, and
  Pechenizkiy]{calders2009building}
T.~Calders, F.~Kamiran, and M.~Pechenizkiy.
\newblock Building classifiers with independency constraints.
\newblock In \emph{Data mining workshops, 2009. ICDMW'09. IEEE international
  conference on}, pages 13--18. IEEE, 2009.

\bibitem[Caruana et~al.(2015)Caruana, Lou, Gehrke, Koch, Sturm, and
  Elhadad]{caruana2015intelligible}
R.~Caruana, Y.~Lou, J.~Gehrke, P.~Koch, M.~Sturm, and N.~Elhadad.
\newblock Intelligible models for healthcare: Predicting pneumonia risk and
  hospital 30-day readmission.
\newblock In \emph{Proceedings of the 21st ACM SIGKDD International Conference
  on Knowledge Discovery and Data Mining}, pages 1721--1730. ACM, 2015.

\bibitem[Chandrasekaran and Jordan(2013)]{chandrasekaran2013computational}
V.~Chandrasekaran and M.~I. Jordan.
\newblock Computational and statistical tradeoffs via convex relaxation.
\newblock \emph{Proceedings of the National Academy of Sciences}, 110\penalty0
  (13):\penalty0 E1181--E1190, 2013.

\bibitem[Cortez et~al.(2009)Cortez, Cerdeira, Almeida, Matos, and
  Reis]{cortez2009modeling}
P.~Cortez, A.~Cerdeira, F.~Almeida, T.~Matos, and J.~Reis.
\newblock Modeling wine preferences by data mining from physicochemical
  properties.
\newblock \emph{Decision Support Systems}, 47\penalty0 (4):\penalty0 547--553,
  2009.

\bibitem[Cotter et~al.(2016)Cotter, Friedlander, Goh, and
  Gupta]{cotter2016satisfying}
A.~Cotter, M.~Friedlander, G.~Goh, and M.~Gupta.
\newblock Satisfying real-world goals with dataset constraints.
\newblock \emph{arXiv preprint arXiv:1606.07558}, 2016.

\bibitem[Dwork et~al.(2012)Dwork, Hardt, Pitassi, Reingold, and
  Zemel]{dwork2012fairness}
C.~Dwork, M.~Hardt, T.~Pitassi, O.~Reingold, and R.~Zemel.
\newblock Fairness through awareness.
\newblock In \emph{Proceedings of the 3rd Innovations in Theoretical Computer
  Science Conference}, pages 214--226. ACM, 2012.

\bibitem[Friedler et~al.(2016)Friedler, Scheidegger, and
  Venkatasubramanian]{friedler2016possibility}
S.~A. Friedler, C.~Scheidegger, and S.~Venkatasubramanian.
\newblock On the (im) possibility of fairness.
\newblock \emph{arXiv preprint arXiv:1609.07236}, 2016.

\bibitem[Goemans and Williamson(1995)]{goemans1995improved}
M.~X. Goemans and D.~P. Williamson.
\newblock Improved approximation algorithms for maximum cut and satisfiability
  problems using semidefinite programming.
\newblock \emph{Journal of the ACM (JACM)}, 42\penalty0 (6):\penalty0
  1115--1145, 1995.

\bibitem[Gouveia et~al.(2013)Gouveia, Parrilo, and Thomas]{gouveia2013lifts}
J.~Gouveia, P.~A. Parrilo, and R.~R. Thomas.
\newblock Lifts of convex sets and cone factorizations.
\newblock \emph{Mathematics of Operations Research}, 38\penalty0 (2):\penalty0
  248--264, 2013.

\bibitem[Gurobi(2016)]{gurobi}
Gurobi.
\newblock Gurobi optimizer reference manual, 2016.
\newblock URL \url{http://www.gurobi.com}.

\bibitem[Hardt et~al.(2016)Hardt, Price, and Srebro]{hardt2016equality}
M.~Hardt, E.~Price, and N.~Srebro.
\newblock Equality of opportunity in supervised learning.
\newblock In \emph{Advances in Neural Information Processing Systems}, pages
  3315--3323, 2016.

\bibitem[Kleinberg et~al.(2016)Kleinberg, Mullainathan, and
  Raghavan]{kleinberg2016inherent}
J.~Kleinberg, S.~Mullainathan, and M.~Raghavan.
\newblock Inherent trade-offs in the fair determination of risk scores.
\newblock \emph{arXiv preprint arXiv:1609.05807}, 2016.

\bibitem[Kocuk et~al.(2016)Kocuk, Dey, and Sun]{kocuk2016inexactness}
B.~Kocuk, S.~S. Dey, and X.~A. Sun.
\newblock Inexactness of sdp relaxation and valid inequalities for optimal
  power flow.
\newblock \emph{IEEE Transactions on Power Systems}, 31\penalty0 (1):\penalty0
  642--651, 2016.

\bibitem[Lasserre(2001)]{lasserre2001global}
J.~B. Lasserre.
\newblock Global optimization with polynomials and the problem of moments.
\newblock \emph{SIAM Journal on Optimization}, 11\penalty0 (3):\penalty0
  796--817, 2001.

\bibitem[Lichman(2013)]{Lichman:2013}
M.~Lichman.
\newblock {UCI} machine learning repository, 2013.
\newblock URL \url{http://archive.ics.uci.edu/ml}.

\bibitem[Lou et~al.(2012)Lou, Caruana, and Gehrke]{lou2012intelligible}
Y.~Lou, R.~Caruana, and J.~Gehrke.
\newblock Intelligible models for classification and regression.
\newblock In \emph{Proceedings of the 18th ACM SIGKDD international conference
  on Knowledge discovery and data mining}, pages 150--158. ACM, 2012.

\bibitem[Louizos et~al.(2015)Louizos, Swersky, Li, Welling, and
  Zemel]{louizos2015variational}
C.~Louizos, K.~Swersky, Y.~Li, M.~Welling, and R.~Zemel.
\newblock The variational fair autoencoder.
\newblock \emph{arXiv preprint arXiv:1511.00830}, 2015.

\bibitem[Luo et~al.(2010)Luo, Ma, So, Ye, and Zhang]{luo2010semidefinite}
Z.-Q. Luo, W.-K. Ma, A.~M.-C. So, Y.~Ye, and S.~Zhang.
\newblock Semidefinite relaxation of quadratic optimization problems.
\newblock \emph{IEEE Signal Processing Magazine}, 27\penalty0 (3):\penalty0
  20--34, 2010.

\bibitem[Madani et~al.(2017)Madani, Sojoudi, Fazelnia, and
  Lavaei]{madani2017finding}
R.~Madani, S.~Sojoudi, G.~Fazelnia, and J.~Lavaei.
\newblock Finding low-rank solutions of sparse linear matrix inequalities using
  convex optimization.
\newblock \emph{SIAM Journal on Optimization}, 27\penalty0 (2):\penalty0
  725--758, 2017.

\bibitem[Miyashiro and Takano(2015)]{miyashiro2015mixed}
R.~Miyashiro and Y.~Takano.
\newblock Mixed integer second-order cone programming formulations for variable
  selection in linear regression.
\newblock \emph{European Journal of Operational Research}, 247\penalty0
  (3):\penalty0 721--731, 2015.

\bibitem[Mosek(2017)]{mosek}
Mosek.
\newblock The {MOSEK} optimization toolbox for {MATLAB} manual, 2017.

\bibitem[Podesta et~al.(2014)Podesta, Pritzker, Moniz, Holdren, and
  Zients]{united2014big}
J.~Podesta, P.~Pritzker, E.~Moniz, J.~Holdren, and J.~Zients.
\newblock \emph{Big data: Seizing opportunities, preserving values}.
\newblock Executive Office of the President, 2014.

\bibitem[Ridgeway et~al.(1998)Ridgeway, Madigan, Richardson, and
  O'Kane]{ridgeway1998interpretable}
G.~Ridgeway, D.~Madigan, T.~Richardson, and J.~O'Kane.
\newblock Interpretable boosted na{\"\i}ve bayes classification.
\newblock In \emph{KDD}, pages 101--104, 1998.

\bibitem[Smola et~al.(2005)Smola, Vishwanathan, and Hofmann]{smola2005kernel}
A.~J. Smola, S.~Vishwanathan, and T.~Hofmann.
\newblock Kernel methods for missing variables.
\newblock In \emph{AISTATS}, 2005.

\bibitem[Tuy(1995)]{tuy1995dc}
H.~Tuy.
\newblock Dc optimization: theory, methods and algorithms.
\newblock In \emph{Handbook of global optimization}, pages 149--216. Springer,
  1995.

\bibitem[Yuille and Rangarajan(2002)]{yuille2002concave}
A.~L. Yuille and A.~Rangarajan.
\newblock The concave-convex procedure (cccp).
\newblock In \emph{Advances in neural information processing systems}, pages
  1033--1040, 2002.

\bibitem[Zafar et~al.(2017)Zafar, Valera, Rodriguez, and Gummadi]{zafar2017}
M.~B. Zafar, I.~Valera, M.~G. Rodriguez, and K.~P. Gummadi.
\newblock Fairness constraints: Mechanisms for fair classification.
\newblock In \emph{Proceedings of the 20th International Conference on
  Artificial Intelligence and Statistics}, 2017.

\bibitem[Zemel et~al.(2013)Zemel, Wu, Swersky, Pitassi, and
  Dwork]{zemel2013learning}
R.~Zemel, Y.~Wu, K.~Swersky, T.~Pitassi, and C.~Dwork.
\newblock Learning fair representations.
\newblock In \emph{Proceedings of the 30th International Conference on Machine
  Learning (ICML-13)}, pages 325--333, 2013.

\bibitem[Zliobaite(2015)]{zliobaite2015relation}
I.~Zliobaite.
\newblock On the relation between accuracy and fairness in binary
  classification.
\newblock \emph{arXiv preprint arXiv:1505.05723}, 2015.

\end{thebibliography}

\end{document}